\documentclass[11pt]{article}
\usepackage[margin=1in]{geometry}
\usepackage{times}
\usepackage{graphicx}
\usepackage{amsmath,amssymb}
\usepackage{booktabs}
\usepackage{multirow}
\usepackage{microtype}
\usepackage{float}
\usepackage{hyperref}
\title{Interactive Whole Slide Images for RL-based Tumour Segmentation}

\author{
    Mohamad Mohamad$^{1}$\thanks{Corresponding author: \texttt{mohamad.mohamad@inria.fr}}
    \quad \quad
    Francesco Ponzio$^{2}$
    \quad \quad
    Maxime Gassier$^{3}$
    \\[0.5em]
    Nicolas Pote$^{3}$
    \quad \quad
    Xavier Descombes$^{1}$
    \\
    \\
    \small
    $^{1}$Université Côte d'Azur, Inria, CNRS, I3S, INSERM, IBV, Sophia Antipolis, France
    \\
     \small
    $^{2}$Department of Control and Computer Engineering,
    Politecnico di Torino, Turin, Italy
    \\
     \small
    $^{3}$Department of Pathology, Bichat Hospital,
    Assistance Publique--Hôpitaux de Paris, Paris, France
}
\date{}

\begin{document}
\maketitle

\begin{abstract}
Whole-slide image (WSI) analysis remains computationally challenging due to the extremely large spatial resolution of slides and the sparse distribution of tumour regions. We propose an end-to-end reinforcement learning framework for sequential tumour segmentation directly on WSIs. Instead of treating the slide as a predefined collection of candidate patches, we formulate the WSI itself as a hierarchical multi-resolution environment through which an agent navigates using movement, zooming, and tumour selection actions. The agent jointly processes local observations and a global thumbnail representation within an actor-critic architecture trained using proximal policy optimization (PPO). Experiments on pulmonary adenocarcinoma WSIs demonstrate the feasibility of direct sequential tumour segmentation on full slides, achieving comparable coarse segmentation quality  relative to patch-based approaches operating at similar magnification levels, while reducing inference time to a few seconds per slide. We further analyse the impact of environment design and action-space granularity. Our results suggest that modelling WSIs as interactive environments provides a promising direction for RL-based computational pathology.
\end{abstract}


\section{Introduction}
\label{sec:intro}

Whole-slide images (WSIs)~\cite{Med} are high-resolution digital scans of histological tissue slides widely used in clinical diagnosis and computational pathology. Due to their extremely large spatial resolution, WSIs are commonly represented as multi-resolution pyramids that enable efficient navigation across magnification levels. In clinical practice, pathologists sequentially explore these pyramidal slides through spatial navigation and zooming, focusing attention on diagnostically relevant tissue regions while avoiding exhaustive inspection of the entire slide.

From a computational perspective, the scale of WSIs makes dense slide-level processing prohibitively expensive. Consequently, most deep learning approaches decompose WSIs into smaller image patches that are processed independently or aggregated across the slide~\cite{transmil}. While effective, such approaches largely discard the sequential and interactive nature of slide exploration, often requiring exhaustive processing of many redundant or diagnostically irrelevant regions.

Reinforcement learning (RL) provides a natural framework for modelling sequential decision-making in interactive visual environments. Recent RL-based approaches for WSI analysis have primarily focused on selecting informative patches from predefined candidate tile sets for downstream classification tasks~\cite{PAMIL,shasha,RLogist}. Earlier methods even constrained the interaction space to pre-extracted image tiles rather than the WSI itself~\cite{first_WSI,breast_WSI}. Although these approaches can improve computational efficiency, they do not directly model the hierarchical WSI representation as an interactive spatial environment. Treating the WSI pyramid itself as the environment would enable clinically faithful and computationally efficient slide navigation, naturally reflecting the coarse-to-fine strategy pathologists employ in practice.

In this work, we investigate end-to-end RL for sequential tumour segmentation directly on full WSIs. We formulate the pyramidal WSI representation itself as a hierarchical multi-resolution environment through which an agent navigates using movement, zooming, and tumour-selection actions. The proposed actor-critic agent jointly processes local observations and a global slide overview while progressively constructing a tumour segmentation mask through sequential interaction with the WSI pyramid. We evaluate the proposed framework on pulmonary adenocarcinoma WSIs and analyse the effects of environment design, action-space granularity, reward formulation, and training strategies. Despite operating under constrained sequential interaction, the proposed agent achieves comparable segmentation performance while substantially improving inference efficiency relative to patch-based baselines.

The main contributions of this work are summarised as follows:

\begin{itemize}
    \item We introduce a reinforcement learning framework for sequential tumour segmentation that models the full WSI pyramid as an interactive multi-resolution environment, and train an end-to-end actor-critic on pulmonary adenocarcinoma WSIs.

    \item We perform extensive analyses of reward design, action-space granularity, architectural choices, and training strategies, providing practical insights into RL-based WSI segmentation.

    \item We compare the proposed framework against patch-level UNI-based classification pipelines across both segmentation performance and inference efficiency.
\end{itemize}

\section{Related Works}
\label{sec:Related Works}

\subsection{RL in medical images}
RL has gained increasing attention in medical image analysis as a principled framework for modelling sequential and spatial decision-making processes~\cite{Survey}. While applications span a broad range of tasks, we focus here on localisation, detection, and segmentation.

Early work formulated anatomical landmark detection as a navigation problem, in which a separate agent per landmark iteratively traverses 2D and 3D volumes via discrete actions, with the reward defined as the change in Euclidean distance to the target~\cite{Landmark1,landmark2,landmark3,3D_landmark}. Extensions of this paradigm introduced a multi-agent setting with shared backbone weights for implicit knowledge sharing~\cite{landmark2} and uncertainty estimation to flag out-of-distribution images for manual review~\cite{landmark3}. This line of work was later extended to organ and lesion localisation~\cite{organs,lesion1}.

Subsequent work explored tighter integration of RL with deep learning architectures. RL-based pre-localisation was combined with UNet-based segmentation for catheter detection~\cite{localisation}, while multi-agent formulations were adopted for iterative and interactive segmentation refinement~\cite{interactive}. RL has also been applied to learn adaptive masking strategies for self-supervised pretraining, improving representation quality and downstream segmentation performance~\cite{masking}.

Segmenting small regions of interest remains challenging when localisation and segmentation are decoupled. A collaborative framework combining RL-based localisation and refinement with UNet-based segmentation was proposed to address this for small targets~\cite{RL-CoSeg}. More recently, segmentation has been formulated as an iterative per-pixel decision process in which pixel labels are progressively refined under a learned policy~\cite{Liu2025}.

\begin{figure}[!t]
\begin{center}
\includegraphics[width=0.9\textwidth]{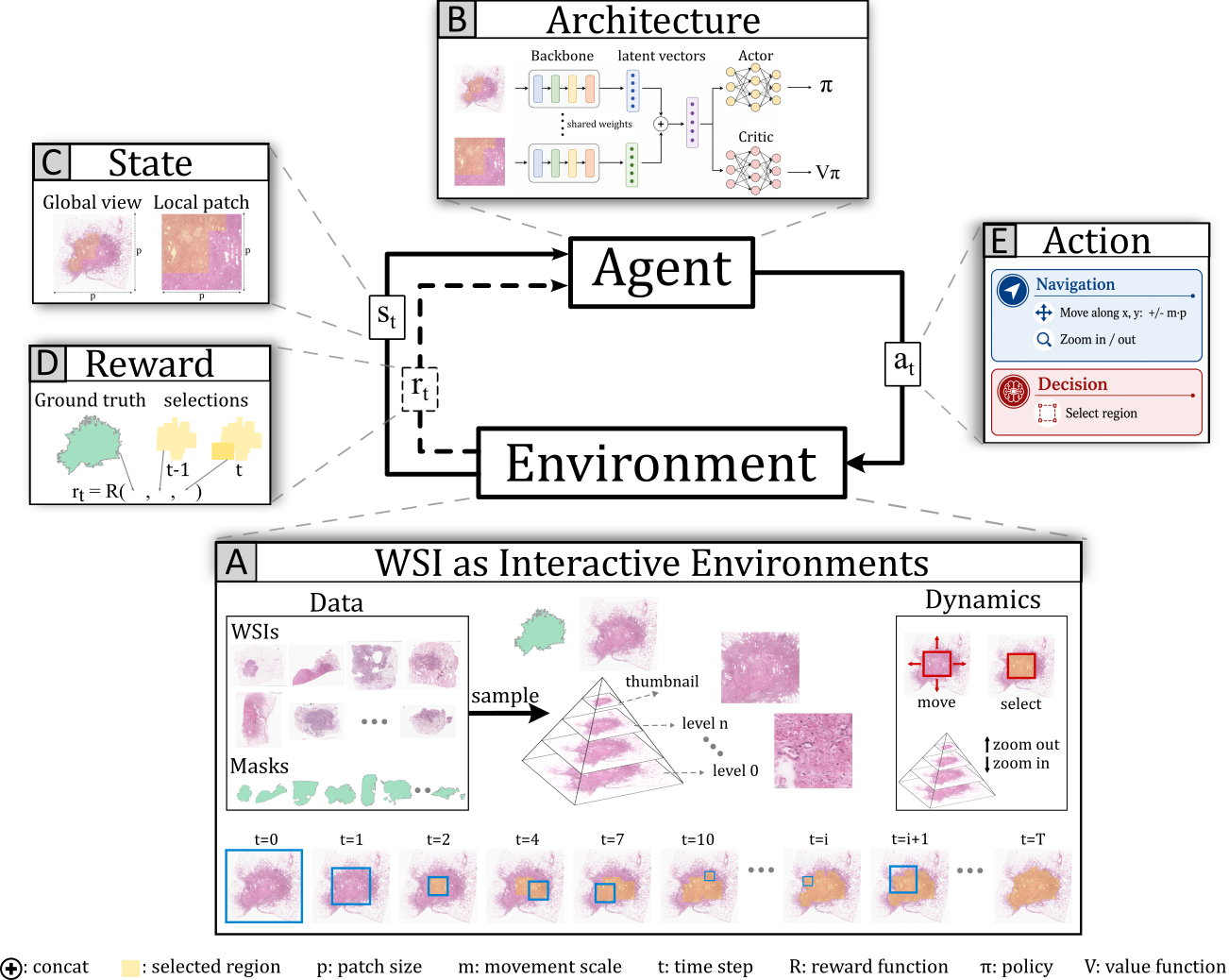}
\end{center}
\caption{Agent framework for sequential WSI tumour segmentation. A) WSIs are formulated as interactive multi-resolution environments. At each episode reset, a WSI and its tumour mask are sampled. The agent sequentially navigates the image pyramid through movement, zooming, and selection actions, progressively constructing an accumulated tumour segmentation mask (yellow). B) Actor--critic architecture jointly processing global and local observations through a shared backbone encoder followed by feature fusion and policy/value heads. C) State representation composed of a global overview and the current local patch. D) Reward computation based on improvements in overlap between accumulated selections and the ground-truth tumour mask. E) Action space composed of navigation and tumour-selection actions.}
\label{fig:pipeline}
\end{figure}

\subsection{RL in Histopathology}

Current work applying RL to histopathological images 
analysis differs in formulation and application, yet shares a common objective: reducing computational cost by selectively attending to diagnostically relevant regions in sparse WSIs.

Early approaches investigated end-to-end trained agents operating directly on image patches~\cite{first_WSI,breast_WSI}. HER2 IHC scoring was formulated as a sequential decision-making problem, where an agent 
iteratively selects a limited number of regions within a tile before producing a final tile-level prediction, with an inhibition-of-return mechanism to discourage repeatedly attending to previously visited regions~\cite{first_WSI}. Similarly, an agent for breast cancer classification on pre-cropped histopathology images was proposed, sequentially aggregating information from selected regions to refine predictions over time~\cite{breast_WSI}.

More recent work has shifted towards pretrained feature extractors and weakly supervised multi-instance learning (MIL) pipelines, where RL primarily serves as a patch selection strategy~\cite{WSI2023,RLogist,PAMIL,shasha}. These methods typically represent WSIs using predefined candidate patches or low-resolution feature maps, from which RL agents sequentially select informative regions. Despite differences in formulation, the learned action space remains largely restricted to spatial patch selection, while zooming and magnification transitions are predefined or externally controlled rather than explicitly learned within the policy itself.

While these approaches improve computational efficiency through selective region processing, they do not directly model the hierarchical WSI pyramid as an interactive environment. More recently, RL environments for histopathological analysis built on top of Gym~\cite{gym} and TorchRL~\cite{torchrl} have been introduced, facilitating the development and evaluation of RL agents for WSI-based tasks~\cite{histogym2024,mohamad}. To the best of our knowledge, our work is the first to formulate WSI 
tumour segmentation as an end-to-end sequential decision-making process operating directly on the full slide pyramid as the environment.

\section{Segmentation Agent and Environment}

We formulate the task as a Markov Decision Process (MDP) defined by the tuple $(\mathcal{S}, \mathcal{A}, \mathcal{R}, \gamma)$, where $\mathcal{S}$ denotes the state space, $\mathcal{A}$ the action space, $\mathcal{R}$ the reward function, and $\gamma \in [0,1]$ the discount factor. At each timestep $t$, the agent observes the current state $s_t$, executes an action $a_t$, and receives a reward $r_t$ from the environment. The discounted return is defined as the cumulative discounted sum of future rewards over an episode:
\begin{equation}
G_t=\sum_{k=t}^{T}\gamma^{k-t} r_k,
\end{equation}
where $T$ denotes the terminal timestep. The objective of the agent is to learn a policy $\pi(a_t \mid s_t)$ that maximises the expected return $\mathbb{E}_\pi[G_t]$ while efficiently identifying diagnostically relevant regions of the WSI.

Similar to earlier end-to-end RL approaches~\cite{breast_WSI,first_WSI}, our method directly trains an agent through sequential interaction with histopathology data. At the same time, like more recent WSI-level methods~\cite{shasha,RLogist,PAMIL}, our framework operates directly on whole-slide images rather than isolated image regions. However, in contrast to prior work, we formulate the WSI itself as the interactive environment, rather than treating it as a predefined set of candidate patches. Specifically, the slide is modelled as a spatial multi-resolution pyramid through which the agent sequentially navigates for tumour segmentation. In the following subsections, we detail the different components of the proposed framework.

\subsection{Environment}
\label{sub:env}
At each environment reset, a WSI together with its corresponding tumour mask is sampled to define the current episode. Let $I$ denote a WSI with corresponding ground-truth mask $M$, and let $p$ denote the fixed observation size. We further denote by $W$ and $H$ the width and height of the slide at the highest-resolution pyramid level ($l=0$).

The environment is modelled as a spatial multi-resolution pyramid composed of levels $\{0,\dots,n\}$ together with a thumbnail representation, from which observations can be sampled. Adjacent pyramid levels differ by a downsampling factor of $2$, such that an observation of size $p\times p$ at level $l$ corresponds to a lower-resolution representation of a $(2p)\times(2p)$ region at level $l-1$. While the observation size remains fixed across all levels, the size of the explorable spatial domain increases across the pyramid. Specifically, the spatial dimensions at level $l$ are defined as:
\begin{equation}
W_l=\frac{W}{2^l}, \qquad H_l=\frac{H}{2^l},
\end{equation}
where the maximum pyramid depth $n$ is chosen such that:
\begin{equation}
W_n \geq p,\quad H_n \geq p,
\qquad\text{while}\qquad
W_{n+1} < p \ \text{ or }\ H_{n+1} < p .
\end{equation}

holds. Consequently, the pyramid depth varies across WSIs and depends on the chosen observation size $p$. The only exception to the fixed $\times2$ scaling occurs between the thumbnail representation and the coarsest pyramid level $n$.

The environment dynamics allow the agent to navigate both spatially and across pyramid levels by zooming in and out. Spatial movement is defined relative to the observation size $p$, where spatial transitions apply horizontal and vertical displacements $(\Delta x,\Delta y)$ scaled by the current observation dimensions. For example, a displacement of $(-1,1)$ shifts the observation window by one patch size left and downward, respectively, while varying movement magnitudes are also permitted. In addition to navigation, the environment maintains an accumulated selection mask corresponding to regions predicted as tumorous by the agent. Both the selection masks and ground-truth masks are propagated across pyramid levels to preserve multi-scale consistency. Figure~\ref{fig:pipeline}A summarises the overall environment dynamics.

\vspace{0.5em}
\noindent
\textbf{State $s_t$.}
At timestep $t$, the state consists of a pair of observations:
\begin{equation}
s_t = (o'_{\text{current}}, o'_{\text{global}}),
\end{equation}

where $o_{\text{current}}$ denotes the current local observation of size $p\times p$ centred at spatial coordinates $(x,y,l)$ within pyramid level $l$, and $o_{\text{global}}$ denotes the thumbnail representation providing global contextual information. Both views are augmented with the accumulated selection mask generated from previous tumour predictions during the episode, producing the masked observations $o'_{\text{current}}$ and $o'_{\text{global}}$ (Figure~\ref{fig:pipeline}C).

\vspace{0.5em}
\noindent
\textbf{Reward $r_t$.}
Several RL approaches for medical image localisation and segmentation employ reward signals based on inter-step improvements of localisation or segmentation metrics~\cite{organs,Liu2025,Landmark1,histogym2024}. However, directly using signed metric differences (e.g. $sign(\Delta (IoU))$) may encourage unstable behaviour or reward exploitation, particularly in tumour segmentation settings where individual selections contribute unevenly to the final segmentation quality.

We therefore define the reward using the positive improvement in accumulated Intersection-over-Union (IoU):

\begin{equation}
r_t=R(s_t,a_t)=\max\left(\Delta(\mathrm{IoU}) \times S^{\mathrm{IoU}_{t-1}},\,0\right),
\end{equation}

where $S \geq 1$ is a scalar controlling the relative importance of late-stage refinement during an episode, and
\begin{equation}
\Delta(\mathrm{IoU}) = \mathrm{IoU}_t - \mathrm{IoU}_{t-1}.
\end{equation}
with $\mathrm{IoU}_t$ denotes the IoU between the ground-truth mask $M$ and the union of all regions selected by the agent up to timestep $t$ (Figure~\ref{fig:pipeline}D). When $S=1$, reward contributions depend solely on the absolute $\mathrm{IoU}$ improvement $\Delta(\mathrm{IoU})$, independently of the segmentation stage. This formulation avoids explicit penalties for incorrect selections while still discouraging them implicitly, since inaccurate selections reduce the maximum achievable $\mathrm{IoU}$ and consequently limit future reward accumulation. In practice, this encourages exploratory behaviour without suppressing selection actions in slides containing small tumour regions. 

The reward is only computed following selection actions, while navigation actions receive zero reward. Reward computation may additionally be performed using lower-resolution pyramid representations of the masks to reduce computational overhead.

\vspace{0.5em}
\noindent
\textbf{Observability and pyramid depth.}
In the current setting, the exploration depth of the pyramid is constrained to the thumbnail representation together with four pyramid levels $\{n,n-1,n-2,n-3\}$. Combined with the global thumbnail observation, this restriction approximately preserves observability of the environment without requiring an explicit memory module. Removing either the global overview or the constrained pyramid depth would instead transform the problem into a partially observable Markov decision process (POMDP), potentially motivating recurrent architectures such as LSTMs similar to~\cite{breast_WSI,first_WSI}. Although recurrent memory mechanisms may further improve performance and are widely used in RL, our primary objective is to evaluate whether long-horizon WSI interaction can generalise across unseen patients without such mechanisms. Incorporating explicit memory modules would introduce an additional layer of architectural and optimisation complexity that falls outside the scope of the present study. We therefore defer its integration to future work.

\subsection{Agent}

The agent architecture consists of three main components: a feature extractor, a feature fusion stage, and separate actor-critic heads. The two image observations composing the state $o'_{\text{current}}$ and $o'_{\text{global}}$ are independently processed using a shared-weight backbone encoder, producing latent feature representations $v_{\text{current}}$ and $v_{\text{global}}$. The resulting feature vectors are concatenated into a joint representation $v$, which is subsequently processed by separate actor and critic heads. The actor predicts a probability distribution over actions, $\pi(.|s_t)$, while the critic estimates the state-value function, $V_\pi(s_t)$. An overview of the architecture is shown in Figure~\ref{fig:pipeline}B.

\vspace{0.5em}
\noindent
\textbf{Actions.}

The action space is divided into two categories: navigation actions and decision actions (Figure~\ref{fig:pipeline}E). Since the agent performs tumour segmentation directly through interaction with the WSI, decision actions correspond to tumour selection over the current observation. In the current formulation, this is represented by a single \textbf{Select} action, which marks the current region as tumourous. Navigation actions consist of spatial movement and zooming operations within the image pyramid. Zooming actions allow transitions across pyramid levels, while spatial actions translate the observation window relative to the current patch size. Let $m$ denote the relative movement factor controlling the translation magnitude. For each value of $m$, four directional actions corresponding to horizontal and vertical movement are defined. We investigate two configurations: $m=0.25$ yielding four directional actions plus zoom-in, zoom-out, and select (\textbf{Agent-7}), and $m \in \{0.1, 0.5, 1\}$ yielding fifteen actions (\textbf{Agent-15}).

\vspace{0.5em}
\noindent
\textbf{Backbone architecture.}
We employ a lightweight ResNet-family backbone~\cite{res} as a feature encoder for both local and global observations, following prior WSI RL approaches based on convolutional patch representations~\cite{PAMIL,RLogist,WSI2023}. While the framework remains backbone-agnostic and extensible to transformer architectures, ResNet encoders provide a favourable trade-off between representational capacity and computational efficiency for long-horizon RL training.

Generalisation in visual RL remains an active research problem~\cite{nethack,CoinRun,Proce}, particularly under limited environment diversity. Since training directly on full WSIs substantially increases complexity while limiting the number of unique training environments, we investigate dropout regularisation as a strategy to improve robustness and mitigate overfitting.

\vspace{0.5em}
\noindent
\textbf{Agent Training.}
Earlier work on WSI analysis trained agents using REINFORCE~\cite{first_WSI,breast_WSI}, while more recent methods~\cite{RLogist,PAMIL,shasha} adopt PPO~\cite{ppo}, which we follow throughout this study. At each timestep, the agent samples an action $a_t \sim \pi_{\theta}(\cdot|s_t)$, receives reward $r_t$, and transitions to the next state $s_{t+1}$. PPO alternates between rollout collection under a fixed policy $\theta_k$ and policy optimisation using an advantage estimate $A_t$. The clipped surrogate objective is defined as:

\begin{equation}
\label{eq:clip}
L^{\text{CLIP}}(\theta) = \mathbb{E}_t\left[ \min\left( \hat{r}_t(\theta) A_t,\; 
\text{clip}(\hat{r}_t(\theta),\, 1-\epsilon,\, 1+\epsilon)\,A_t \right) \right]
\end{equation}

where $\hat{r}_t(\theta) = \pi_\theta(a_t|s_t) / \pi_{\theta_k}(a_t|s_t)$ denotes the probability ratio between the current and previous policy. The full PPO objective additionally incorporates a value loss and entropy bonus:

\begin{equation}
\label{eq:ppo}
L^{\text{PPO}}(\theta) = \mathbb{E}_t\left[ 
L^{\text{CLIP}}_t(\theta) 
- c_1 \left( V_w(s_t) - V_t^{\text{target}} \right)^2 
+ c_2\, \mathcal{H}[\pi_\theta](s_t) 
\right]
\end{equation}

where $V_w$ denotes the critic, $\mathcal{H}$ the entropy bonus, and $c_1, c_2$ weighting coefficients.

Incorporating batch normalisation and dropout into RL training can introduce optimisation instabilities due to discrepancies between training and evaluation behaviour~\cite{me,CrossQ,CrossNorm}. Prior WSI RL approaches~\cite{WSI2023,RLogist,PAMIL} largely avoid such issues by freezing the backbone and using it solely as a feature extractor. In contrast, our backbone is trained jointly with the agent, motivating the use of MDR~\cite{me}, a recently proposed method for stable RL optimisation under batch normalisation and dropout regularisation. We compare MDR against standard PPO training in the ablation study (Section~\ref{sub:mdr}).

\section{Experiments}

We conduct ablation studies on the main components of the proposed agent and environment. In particular, we analyse the effect of the reward scaling parameter $S$ (Section~\ref{sub:reward-ablation}), architectural choices related to dropout and MDR-based training stabilisation (Section~\ref{sub:mdr}), and the impact of episode horizon $T$ and action-space granularity $m$ (Section~\ref{sub:action-horizon}). We further compare the proposed framework against a patch-level classification baseline in terms of both segmentation performance and inference efficiency on the evaluation cohort (Section~\ref{sub:result}).

Unless otherwise specified, all experiments use a ResNet18 backbone, the Agent-7 configuration, $S=4$, max timesteps $T=80$, and an observation size of $p=112$. All experiments are conducted across three random seeds. Additional implementation details and hyperparameter settings are provided in Appendix~\ref{appendix:implementation}.

\begin{figure}[!t]
\begin{center}
\includegraphics[width=0.99\textwidth]{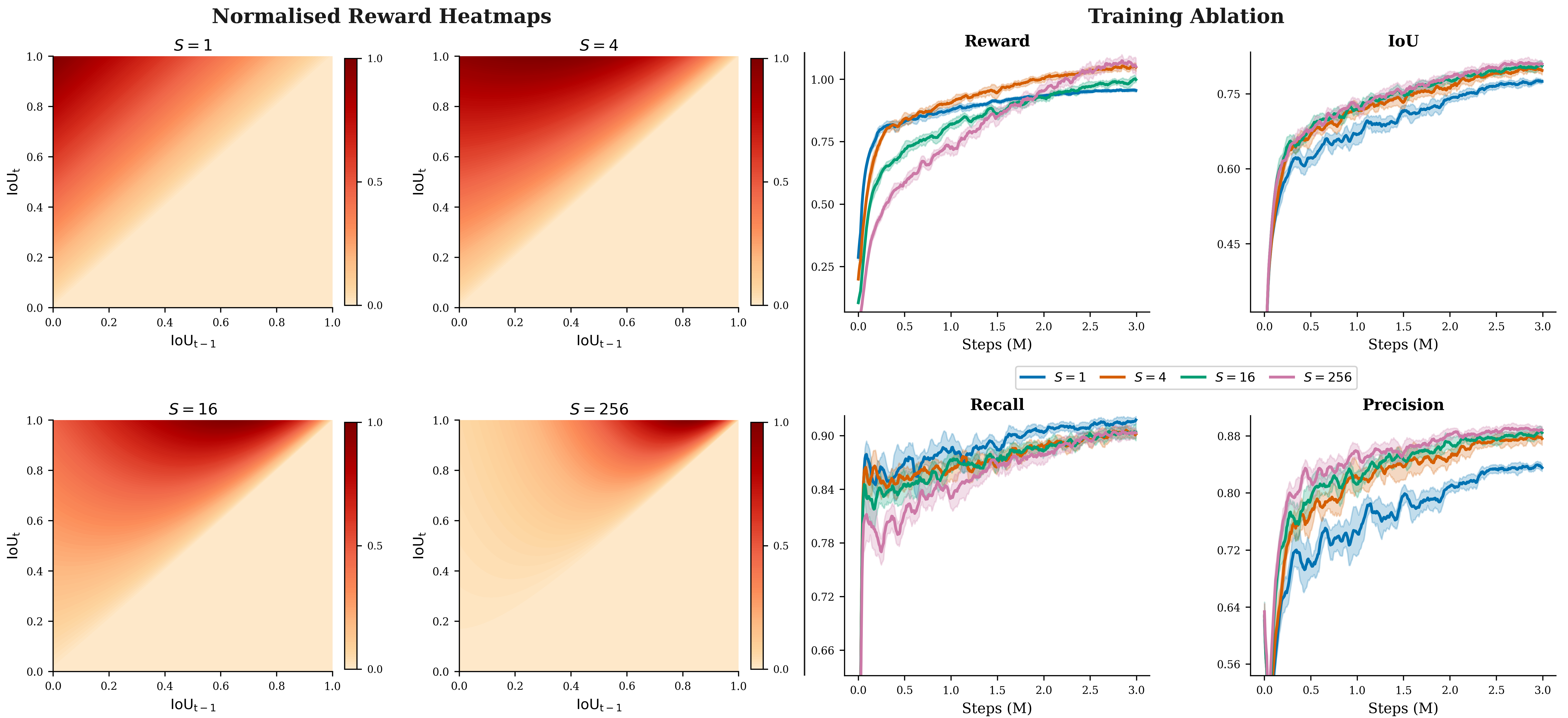}
\end{center}

\caption{Reward shaping analysis. Left: Normalised reward landscapes for different scaling parameters $S$ as a function of consecutive IoU values $(\mathrm{IoU}_{t-1}, \mathrm{IoU}_t)$. Increasing $S$ shifts the reward towards improvements obtained at higher IoU regimes, encouraging late-stage refinement. Right: Training dynamics for different values of $S$. Lower values favour broader region selection and higher recall early during training, whereas larger values promote more conservative refinement with improved precision.}

\label{fig:reward_ablation}
\end{figure}

\subsection{Dataset}
\label{sub:data}
Experiments were conducted on a pulmonary adenocarcinoma dataset collected at H\^opital Pasteur for tumour segmentation. The original cohort consisted of 94 patients with one or more WSIs, resulting in 106 annotated slides. The cohort is balanced across sex, with a mean patient age of 65.

Each slide was annotated independently by two pathologists, one junior and one senior, using QuPath. Tumour regions were delineated precisely so as to exclude artefacts, non-tumoural areas, and adjacent tissue such as parenchyma and extratumoural stroma. The resulting tumour regions vary considerably in size, occupying $23.7 \pm 11.3\%$ of the whole slide and $45.3 \pm 19.3\%$ of the tissue area on average. This variability spans a wide range: measured as a fraction of the whole slide, 9 slides contain less than $10\%$ tumour, 62 less than $25\%$, and 104 less than $50\%$; measured as a fraction of tissue area, the corresponding counts are 2, 15, and 67 slides.

The dataset was split at the patient level into training and test cohorts, with 65 patients (71 WSIs) used for training and 29 patients (35 WSIs) reserved for testing. We refer to this subset as \textit{Original-Cohort}. To increase the number of environments available during RL training, we further extended the training cohort by incorporating additional WSIs from the same training patients, increasing the number of training slides from 71 to 334 (+263 WSIs). Since manual annotations were unavailable for these additional slides, tumour masks were generated automatically using UNI-ENS, a patch-level ensemble baseline described in Section~\ref{sub:result}, and were used exclusively during training.

Evaluation is performed only on the pathologist-annotated test slides of \textit{Original-Cohort}, ensuring that pseudo-labels influence the training set alone and not the validity of reported results. Unless otherwise specified, ablation studies use \textit{Original-Cohort}, while final results are obtained using the extended training set.

\begin{figure}[!t]
\begin{center}
\includegraphics[width=0.99\textwidth]{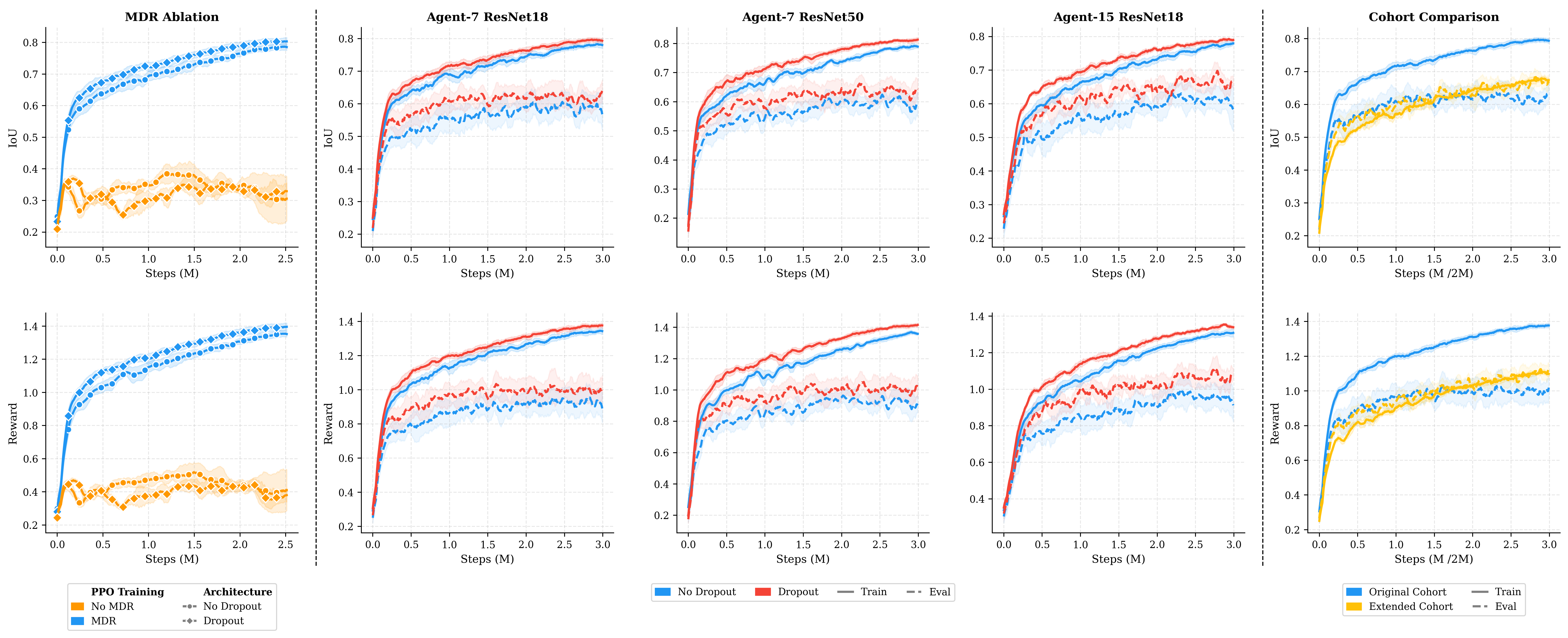}
\end{center}
\caption{Training stability and generalisation analysis. Left: MDR stabilises PPO training and substantially improves convergence and segmentation performance compared to standard PPO. Middle: Effect of dropout regularisation across different agent and backbone configurations. Dropout improves evaluation performance, particularly for higher-capacity agents, although a noticeable train--test gap remains. Right: Extending the training cohort substantially reduces the generalisation gap and improves evaluation performance. Training steps are doubled for agents trained on the extended cohort.}

\label{fig:mdr_gen}
\end{figure}
\subsection{Reward Ablation}
\label{sub:reward-ablation}

The reward formulation introduced in Section~\ref{sub:env} uses improvements in IoU as the learning signal while introducing a scaling parameter $S$ to modulate the contribution of late-stage segmentation refinement. To analyse the effect of this parameter on both optimisation and learned behaviour, we evaluate the agent using four values of $S \in \{1,4,16,256\}$.

The main distinction lies between the case $S=1$ and larger values of $S$. When $S=1$, rewards depend solely on the absolute IoU improvement at each step. In contrast, larger values increasingly favour improvements obtained at higher existing IoU values, thereby emphasising late-stage refinement during an episode. Although the reward remains globally maximised at a final IoU of 1 independently of $S$, the resulting navigation and selection strategies differ (see Appendix~\ref{appendix:reward} for additional discussion on learned behaviour).

Figure~\ref{fig:reward_ablation} illustrates the effect of varying $S$. Increasing $S$ shifts higher reward regions towards improvements occurring at already high IoU values. The corresponding training curves reveal a clear precision--recall trade-off: lower values of $S$ initially favour broader recall-oriented behaviour, whereas larger values encourage more conservative selection strategies with higher precision. Intuitively, imprecise selections reduce the maximum achievable IoU during subsequent refinement steps, thereby favouring precision over recall when late-stage improvements are strongly rewarded. In contrast, low recall remains recoverable through future selections and therefore imposes a weaker constraint on long-term reward accumulation.

Very large values of $S$ may additionally produce disproportionately high rewards, which can negatively affect optimisation stability. Overall, the IoU and precision trends suggest that moderate values of $S>1$ (e.g., S=4 or S=16) provide a better optimisation trade-off, encouraging refinement-oriented behaviour while maintaining stable training dynamics.

\subsection{Generalization and MDR}
\label{sub:mdr}

Figure~\ref{fig:mdr_gen} (left) illustrates the effect of MDR on training stability. Without MDR, PPO training of batch-normalisation-based architectures exhibits severe instability, with IoU stagnating near $0.3$ throughout training. Incorporating MDR resolves this instability, establishing MDR as an essential component for jointly training architectures containing batch normalisation and dropout within the proposed RL framework.

Figure~\ref{fig:mdr_gen} (middle) shows the effect of dropout regularisation across agent and backbone configurations. Dropout consistently reduces the train--evaluation gap and improves evaluation IoU across all three configurations, with the benefit appearing more pronounced for higher-capacity agents. Nevertheless, a substantial generalisation gap remains when training on the limited \textit{Original-Cohort}, motivating the cohort extension.

Figure~\ref{fig:mdr_gen} (right) shows that extending the training cohort substantially narrows the train--evaluation gap across both IoU and reward, yielding improvements that exceed those from dropout regularisation alone. Agents trained on the extended cohort require longer training horizons (doubled step budget) but converge to notably higher evaluation performance. These results highlight the importance of environmental diversity for training generalisable RL agents directly on full WSIs.

\begin{table}[t]
\caption{Effect of episode horizon $T$ and action-space granularity on segmentation performance. Performance improves from short to moderate horizons before saturating, while Agent-15 largely benefits the low-budget setting.}
\label{tab:T}
\centering
\small
\setlength{\tabcolsep}{5pt}
\begin{tabular}{|c|ccc|ccc|}
\hline
\multirow{2}{*}{$T$} & \multicolumn{3}{c|}{Agent-7} & \multicolumn{3}{c|}{Agent-15} \\
\cline{2-7}
 & Dice$\uparrow$ & NSD@512$\uparrow$ & NSD@2048$\uparrow$ 
 & Dice$\uparrow$ & NSD@512$\uparrow$ & NSD@2048$\uparrow$ \\
\hline\hline
20 & 0.842 & 0.147 & 0.484 & 0.873 & 0.179 & 0.555 \\
40 & 0.887 & 0.188 & 0.585 & 0.890 & 0.206 & 0.609 \\
60 & 0.893 & 0.201 & 0.612 & 0.887 & 0.209 & 0.612 \\
80 & 0.894 & 0.206 & 0.617 & 0.888 & 0.209 & 0.613 \\
\hline
\end{tabular}
\end{table}

\subsection{Action-space and horizon analysis}
\label{sub:action-horizon}
We ablate agent performance on the training cohort with respect to the maximum episode horizon $T$. An episode terminates either when 95\% of the tumour area has been selected or when the maximum number of allowed steps is reached. Table~\ref{tab:T} reports the performance of Agent-7 and Agent-15 for $T \in \{20,40,60,80\}$ using Dice and Normalized Surface Dice (NSD), where NSD measures boundary agreement between predicted and ground-truth masks within a fixed tolerance (reported at 512 and 2048 pixels); standard deviations are omitted for readability.

Performance consistently decreases when limiting the episode horizon to $T=20$, indicating that short trajectories restrict the agent's ability to refine tumour regions. In contrast, evaluation performance largely plateaus for larger values of $T$, suggesting that many slides can already be segmented within fewer than 60 interaction steps and that later selections contribute primarily to refinement.

The performance drop at low horizons is less pronounced for Agent-15 than for Agent-7, likely due to its larger and more flexible action space, which enables broader spatial exploration within fewer interaction steps. At longer horizons, Agent-15 performs comparably to Agent-7, suggesting that the larger action space may require additional training to exploit the extended step budget. Notably, NSD metrics improve more substantially than Dice with increasing $T$, suggesting that additional steps primarily contribute to boundary refinement rather than coarse tumour coverage. Additional qualitative comparisons of the learned navigation strategies are provided in Appendix~\ref{appendix:steps}.

\subsection{Results}
\label{sub:result}
\begin{table}[t]
\caption{Segmentation performance on the evaluation set. ± denotes standard deviation across three seeds.}
\label{tab:results}
\centering
\small
\setlength{\tabcolsep}{4pt}
\begin{tabular}{lccccc}
\hline
Method & Dice$\uparrow$ & Recall$\uparrow$ & Precision$\uparrow$ & NSD@512$\uparrow$ & NSD@2048$\uparrow$ \\
\hline

\multicolumn{6}{c}{\textit{Patch-based methods}} \\
\hline
UNI-1024 & 0.88 {\scriptsize$\pm$ .004} & 0.85 {\scriptsize$\pm$ .030} & 0.91 {\scriptsize$\pm$ .030} & 0.76 {\scriptsize$\pm$ .041} & 0.88 {\scriptsize$\pm$ .046} \\
UNI-2048 & 0.84 {\scriptsize$\pm$ .025} & 0.96 {\scriptsize$\pm$ .007} & 0.75 {\scriptsize$\pm$ .044} & 0.59 {\scriptsize$\pm$ .027} & 0.74 {\scriptsize$\pm$ .032} \\
UNI-4096 & 0.81 {\scriptsize$\pm$ .034} & 0.93 {\scriptsize$\pm$ .060} & 0.74 {\scriptsize$\pm$ .100} & 0.61 {\scriptsize$\pm$ .009} & 0.74 {\scriptsize$\pm$ .066} \\
UNI-ENS  & 0.90 {\scriptsize$\pm$ .015} & 0.93 {\scriptsize$\pm$ .006} & 0.87 {\scriptsize$\pm$ .030} & 0.91 {\scriptsize$\pm$ .036} & 0.91 {\scriptsize$\pm$ .036} \\
UNI-HR  & 0.89 {\scriptsize$\pm$ .008} & 0.92 {\scriptsize$\pm$ .011} & 0.88 {\scriptsize$\pm$ .024}  & 0.71 {\scriptsize$\pm$ .030} & 0.85 {\scriptsize$\pm$ .035} \\
\hline
\multicolumn{6}{c}{\textit{Sequential methods}} \\
\hline
Agent-7-Res18 & 0.80 {\scriptsize$\pm$ .004} & 0.87 {\scriptsize$\pm$ .018} & 0.76 {\scriptsize$\pm$ .014} & 0.14 {\scriptsize$\pm$ .005} & 0.44 {\scriptsize$\pm$ .010} \\

Agent-7-Res50 & 0.83 {\scriptsize$\pm$ .003} & 0.89 {\scriptsize$\pm$ .005} & 0.81 {\scriptsize$\pm$ .006} & 0.16 {\scriptsize$\pm$ .003} & 0.48 {\scriptsize$\pm$ .039} \\

Agent-15-Res50 & 0.84 {\scriptsize$\pm$ .005} & 0.88 {\scriptsize$\pm$ .002} & 0.82 {\scriptsize$\pm$ .007} & 0.19 {\scriptsize$\pm$ .003} & 0.56 {\scriptsize$\pm$ .008} \\

Human-7 & 0.91 & 0.86 & 0.96 & 0.26 & 0.72 \\
\hline
\end{tabular}
\end{table}

We compare the proposed RL agent against patch-based deep learning baselines built on UNI~\cite{chen2024uni}, a self-supervised foundation model pretrained on over 100,000 histopathology WSIs. Three classifiers are trained using patches corresponding to effective spatial regions of $1024\times1024$, $2048\times2048$, and $4096\times4096$ pixels at the highest slide magnification, resized to $224\times224$. These models are denoted UNI-1024, UNI-2048, and UNI-4096, respectively. Final segmentation masks are obtained through patch aggregation followed by connected-component filtering. We additionally report an ensemble model (UNI-ENS) obtained through patch-wise majority voting across the three classifiers.

We additionally evaluate a hierarchical coarse-to-fine baseline (UNI-HR) that combines the UNI-4096 and UNI-1024 models. A coarse segmentation is first obtained with UNI-4096, after which the predicted boundary is used to identify locations for high-resolution refinement with UNI-1024. Four $1024\times1024$ patches are sampled on each side of the predicted boundary, spanning the inner and outer boundary regions. Only non-overlapping patches are extracted and fed to UNI-1024. The resulting fine-scale predictions are then used to refine the coarse segmentation mask.

The RL agents are trained on the extended cohort, while evaluation is performed exclusively on the pathologist-annotated evaluation cohort. We report three agent configurations spanning two backbones and two action spaces. Agent-7-Res18 and Agent-7-Res50 share the seven-action space but use ResNet18 and ResNet50 backbones, respectively, allowing assessment of the effect of backbone capacity. Agent-15-Res50 retains the ResNet50 backbone while expanding the action space to fifteen actions, allowing assessment of the effect of finer action granularity. Implementation details are provided in Appendix~\ref{appendix:implementation}. We further include a constrained human baseline (Human-7), where the annotator interacts with the slide using the same seven-action space and episode horizon as the agent while having access to the ground-truth segmentation mask.

Table~\ref{tab:results} reports segmentation performance across Dice, Recall, Precision, and NSD metrics. Among the UNI-based approaches, UNI-1024 achieves the strongest single-model performance across Dice and NSD metrics, while UNI-ENS obtains the best overall performance. The hierarchical UNI-HR baseline reaches a Dice score comparable to UNI-ENS and improves boundary quality over its UNI-4096 starting point, with NSD@512 increasing from $0.61$ to $0.71$ and NSD@2048 from $0.74$ to $0.85$. This confirms that explicit high-magnification refinement can improve boundary agreement, although UNI-HR does not match UNI-ENS or UNI-1024 in boundary quality.

Among the sequential methods, Agent-7-Res18 and Agent-7-Res50 achieve Dice scores comparable to UNI-4096 and UNI-2048, respectively, despite operating under substantially stronger interaction constraints. Agent-15-Res50 is the strongest sequential agent across all reported metrics, achieving a Dice score of $0.84$ and an NSD@2048 of $0.56$. Its improvement over the seven-action agents suggests that the finer action space can provide a measurable benefit when combined with a sufficient training budget. Since the agents are restricted to navigating only the upper levels of the WSI pyramid, their explored spatial scale is most comparable to UNI-4096.

\begin{figure}[!t]
\begin{center}
\includegraphics[width=0.98\textwidth]{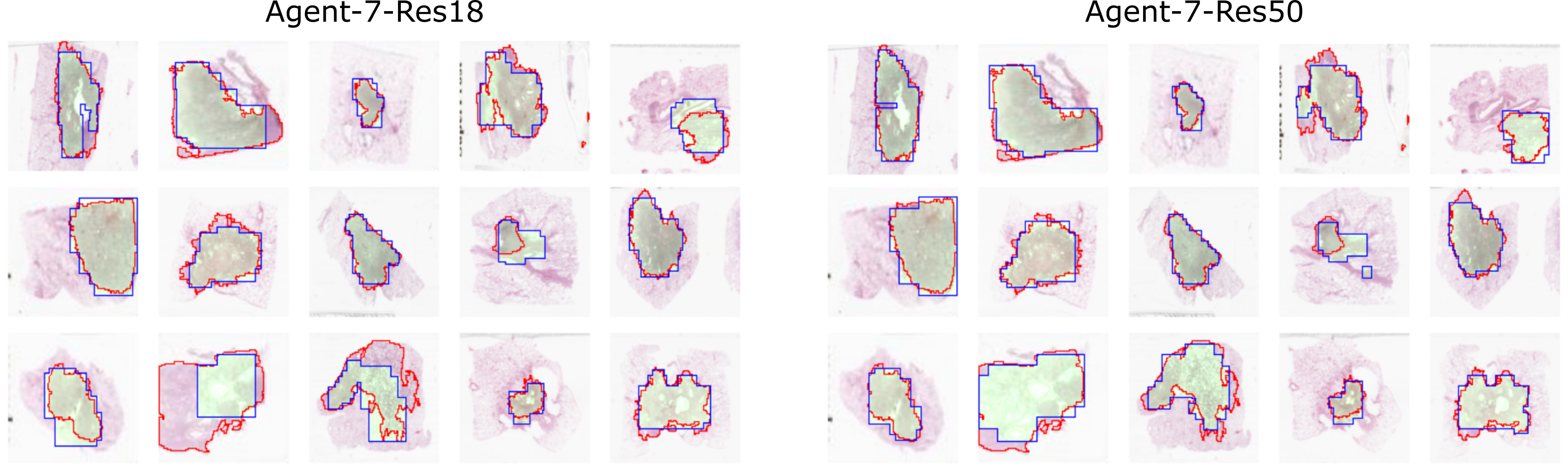}
\end{center}
\caption{Segmentation results on evaluation slides. Predictions from Agent-7 ResNet18 (left) and Agent-7 ResNet50 (right) are shown for tumours of varying size and morphology. Red contours denote the ground-truth tumour boundaries, while the blue contours correspond to the final agent segmentation obtained from the accumulated selections at the end of the episode.}
\label{fig:eval_eps}
\end{figure}

The largest performance gap appears in NSD metrics, indicating that most agent errors occur near tumour boundaries. This behaviour is expected given the coarse spatial granularity imposed by patch-level selection actions and the limited depth of the WSI pyramid. Nevertheless, the improvement from NSD@512 to NSD@2048 suggests that many errors remain spatially close to the true boundaries rather than corresponding to large localisation failures. This observation is further supported by Figure~\ref{fig:eval_eps}, where most discrepancies between the predicted and ground-truth masks remain localised near tumour boundaries. Notably, the constrained human annotator, despite having access to the ground-truth mask, achieves an NSD@512 of only $0.26$ under the same seven-action constraints, increasing substantially to $0.72$ at NSD@2048. This mirrors the agents' behaviour and suggests that the boundary gap is predominantly structural, imposed by the coarse action space and pyramid constraints, rather than reflecting a fundamental limitation of the learned policy.

\begin{table}
\caption{Inference time per slide on the evaluation set. The proposed RL agent substantially improves efficiency. ± is across slides and seeds (Mean) or across seeds (Best/Worst-case, computed per-seed).}
\label{tab:runtime}
\centering
\small
\setlength{\tabcolsep}{5pt}
\begin{tabular}{lccc}
\hline
Method & Mean Time (s)$\downarrow$ & Worst-case (s)$\downarrow$ & Best-case (s)$\downarrow$ \\
\hline

\multicolumn{4}{c}{\textit{Patch-based methods}} \\
\hline
UNI-1024 & 229.1 {\scriptsize$\pm$ 164.4} & 829.1 {\scriptsize$\pm$ 266.6} & 79.6 {\scriptsize$\pm$ 1.7} \\
UNI-2048 & 187.9 {\scriptsize$\pm$ 118.1} & 538.3 {\scriptsize$\pm$ 10.4} & 64.1 {\scriptsize$\pm$ 2.9} \\
UNI-4096 & 36.8 {\scriptsize$\pm$ 40.2} & 138.7 {\scriptsize$\pm$ 3.0} & 10.0 {\scriptsize$\pm$ 0.4} \\

UNI-HR & 130 {\scriptsize$\pm$ 110.2} & 433 {\scriptsize$\pm$ 14.7} & 36.6 {\scriptsize$\pm$ 1.6} \\

\hline
\multicolumn{4}{c}{\textit{Sequential methods}} \\
\hline
Agent-7-Res18 & \textbf{3.0 {\scriptsize$\pm$ 1.5}} &  \textbf{6.0 {\scriptsize$\pm$ 1.3}} &  \textbf{1.0 {\scriptsize$\pm$ 0.2}} \\

Agent-7-Res50 & \textbf{3.6 {\scriptsize$\pm$ 1.5}} &  \textbf{6.5 {\scriptsize$\pm$ 0.1}} &  \textbf{1.2 {\scriptsize$\pm$ 0.1}} \\

Agent-15-Res50 & \textbf{3.1 {\scriptsize$\pm$ 0.7}} &  \textbf{5.1 {\scriptsize$\pm$ 0.6}} &  \textbf{1.8 {\scriptsize$\pm$ 0.1}} \\

\hline
\end{tabular}
\end{table}
We additionally evaluate inference efficiency on the evaluation cohort. Table~\ref{tab:runtime} reports average, best-case, and worst-case processing times per slide (Appendix~\ref{appendix:time}). For UNI-based methods, extraction is optimised by directly sampling patches from appropriate magnification levels and parallelising CPU tiling with GPU inference. Classification, mask aggregation, and post-processing times are omitted as they contribute minimally relative to patch extraction.

Processing larger spatial regions substantially reduces inference time for UNI-based models by decreasing the number of extracted patches. However, processing time remains highly dependent on slide size, leading to large discrepancies between average and worst-case runtimes. Even UNI-4096 requires more than 30 seconds on average and exceeds two minutes in the worst case. The hierarchical UNI-HR baseline is slower still, averaging 130 seconds per slide and exceeding seven minutes in the worst case, since its boundary refinement stage adds a second round of high-magnification patch extraction. In contrast, RL agents process slides within 3 seconds while exhibiting substantially lower runtime variance across slides (worst case at 6s and best at 1s). Relative to UNI-1024, this corresponds to mean speedups of approximately $64$--$76\times$ across the three agent configurations, with a Dice reduction of $0.05$--$0.08$.

Overall, the proposed RL formulation demonstrates that direct sequential tumour segmentation on full WSIs is feasible under constrained interaction settings. While the agents remain less accurate than patch-based UNI approaches, particularly around fine tumour boundaries, they achieve substantially lower inference times and more consistent processing costs across slides.

\section{Discussion}
The results demonstrate the feasibility of direct sequential tumour segmentation on full WSIs under real-time inference constraints, while highlighting how design choices influence the trade-off between segmentation quality and computational efficiency. Improved generalisation under increased training environment diversity further suggests that scaling the number of training environments is a more impactful direction than architectural regularisation alone.

Several limitations remain. The current environment constrains exploration to four pyramid levels and a thumbnail, limiting access to finer multi-scale information. The agents operate without explicit memory mechanisms, restricting information aggregation across 
observations and magnification levels during long-horizon interaction. Finally, the study is monocentric and limited to pulmonary adenocarcinoma, although the proposed formulation is applicable to other organs, tumour types, and clinical centres.

These limitations point to concrete next steps. At the agent level, incorporating explicit memory mechanisms would enable cross-location and cross-scale context aggregation during long-horizon interaction. At the environment level, removing the pyramid-depth constraint would move towards fully unconstrained slide navigation. At the evaluation level, extending beyond a single centre and tumour type would test the generalisability of the formulation across organs and clinical sites. Finally, on the training side, scaling through higher-capacity and transformer-based backbones, pathology foundation models such as UNI~\cite{chen2024uni}, larger datasets, and recent advances in self-supervised and contrastive RL~\cite{contrastive} represent promising avenues for learning more robust navigation directly from large-scale WSI environments.

\section{Conclusion}
We presented an end-to-end reinforcement learning framework for sequential tumour segmentation directly on whole-slide images modelled as interactive multi-resolution environments. The proposed formulation enables agents to navigate WSIs through spatial movement and zooming while performing segmentation under constrained interaction budgets. Experiments on pulmonary adenocarcinoma WSIs demonstrate the feasibility of direct sequential segmentation on full slides, achieving comparable coarse segmentation performance together with substantially improved inference efficiency relative to patch-based baselines. Beyond segmentation performance, our study provides insights into the role of reward design, action-space granularity, optimisation stability, and environment diversity in RL-based computational pathology.

\section{Acknowledgement}
This work has been supported by the French government, through the France 2030 investment plan managed by the Agence Nationale de la Recherche, as part of the ``UCA DS4H'' project, reference ANR-17-EURE-0004.

\bibliographystyle{plain}
\bibliography{egbib}

\appendix

\section{Reward-Induced Behaviour}
\label{appendix:reward}

As discussed in Section~\ref{sub:reward-ablation}, $S>1$ scales rewards towards late-stage IoU improvements, making precise delimitation more valuable than early coarse coverage and directly shaping the selection strategy learned by the agent, since each selection affects the maximum achievable reward for all future steps.

\begin{figure}[!htbp]
\begin{center}
\includegraphics[width=0.95\textwidth]{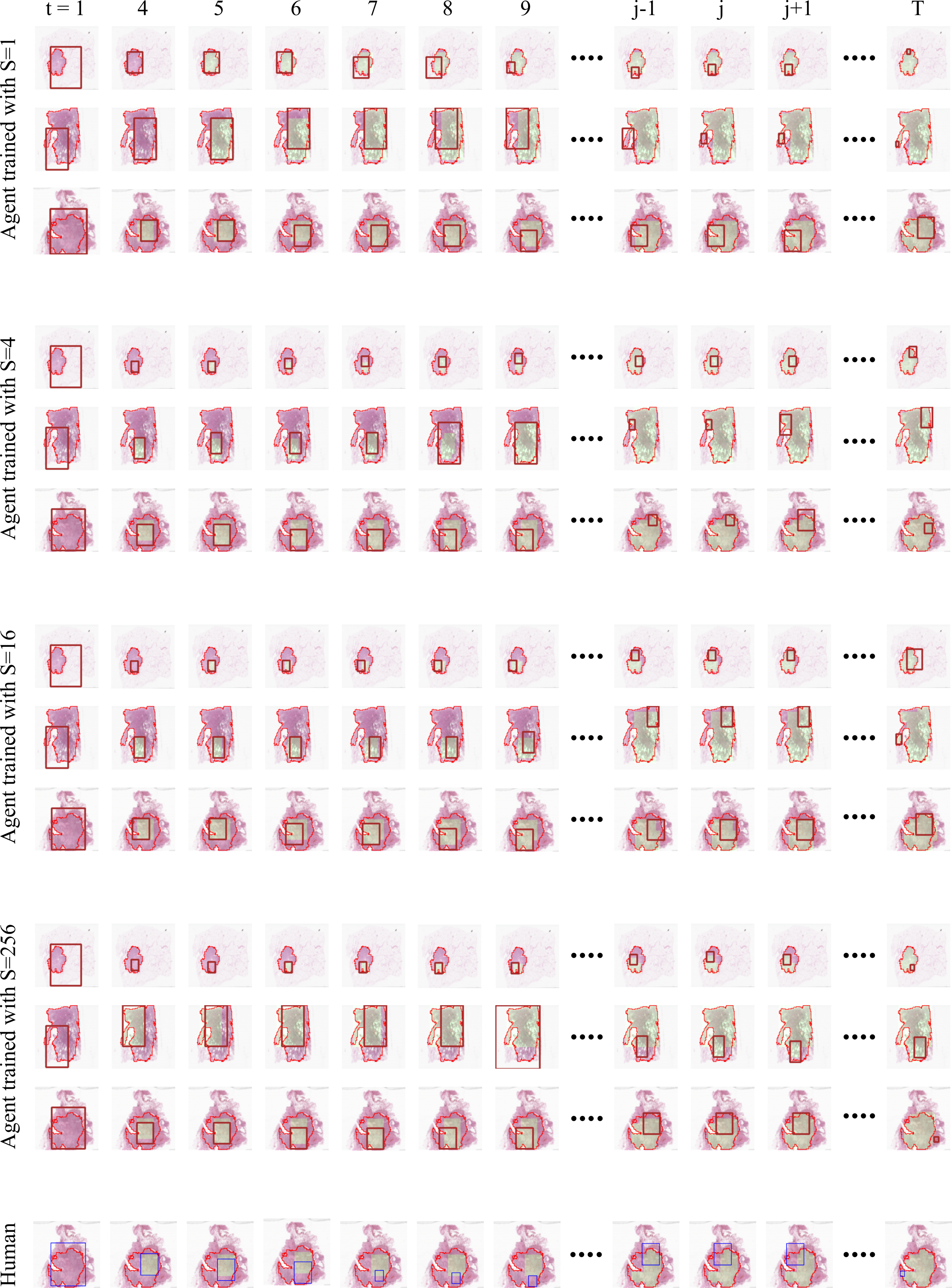}
\end{center}
\caption{Navigation strategies learned under different reward scaling values $S \in \{1,4,16,256\}$ together with the constrained human strategy shown on a training-cohort slides. Episode progression is shown from left to right, with the final timestep corresponding to the end of the episode.}
\label{fig:reward_eps}
\end{figure}

\begin{figure}[!t]
\begin{center}
\includegraphics[width=0.95\textwidth]{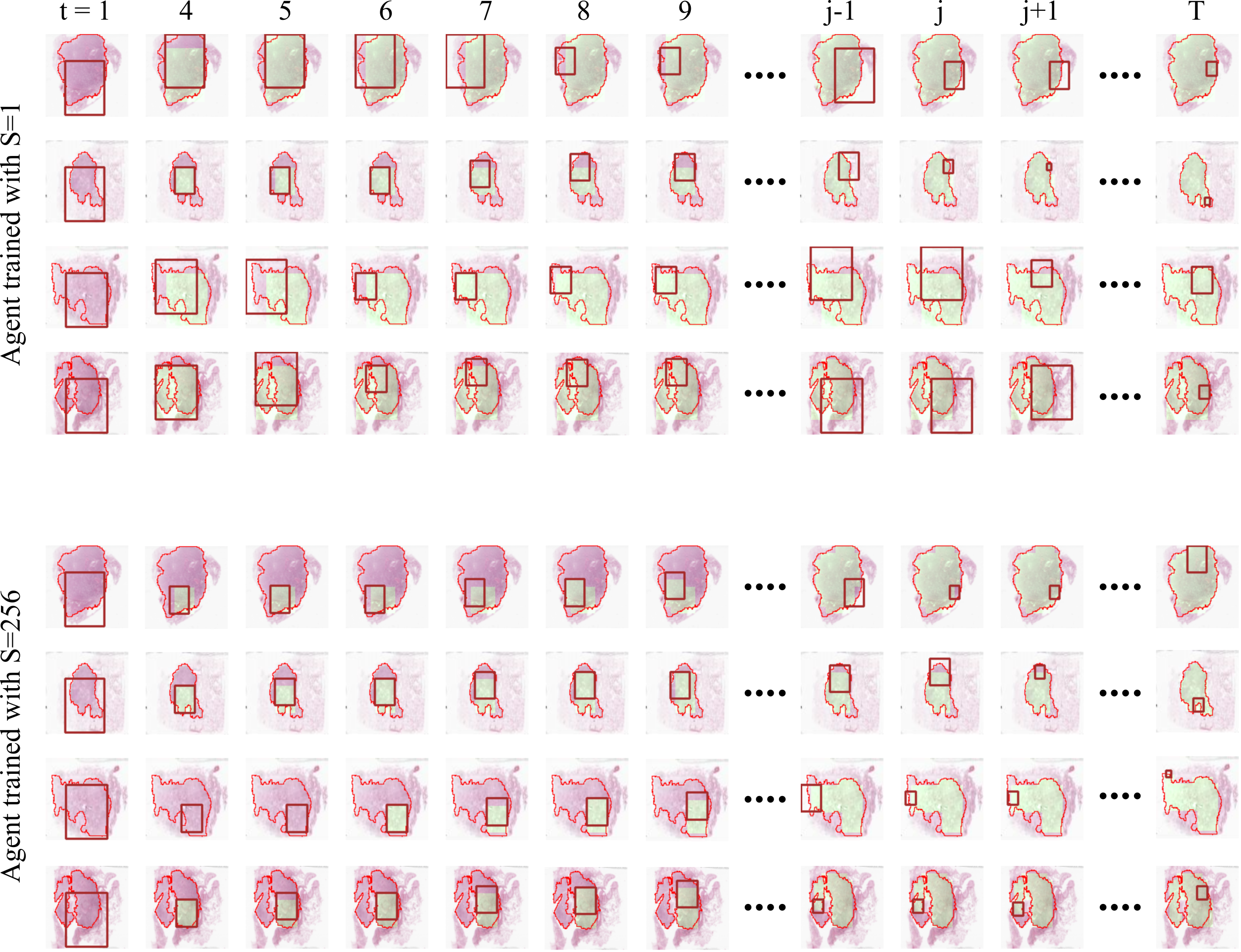}
\end{center}
\caption{Additional navigation episodes for agents trained with $S \in \{1,256\}$.}
\label{fig:reward_eps2}
\end{figure}

Figures~\ref{fig:reward_eps} and~\ref{fig:reward_eps2} show representative episodes for agents trained under different values of $S$ compared against the constrained human strategy. The behavioural distinction is especially pronounced in Figure~\ref{fig:reward_eps2}, where irregularly shaped tumours are present.

Agents trained with $S=1$ exhibit high tolerance to precision sacrifice, favouring early selection of large boxes that rapidly cover substantial tumour area even at the cost of including significant non-tumour regions. This maximises short-term recall but accumulates precision errors that limit future refinement quality. Agents trained with $S>1$ show substantially lower tolerance to such precision sacrifice — they allocate a larger fraction of the step budget to traversing tumour boundaries through multiple smaller selections before committing to broader coverage, trading coverage speed for delimitation quality.

This contrast is most visible for $S=256$ in Figure~\ref{fig:reward_eps2}, where the agent spends a substantial portion of its budget traversing the irregular tumour boundary before expanding coverage, such trajectories resemble the constrained human strategy, particularly during early episode stages.

\section{Episode Horizon and Learned Behaviour}

\begin{figure}[!t]
\begin{center}
\includegraphics[width=0.95\textwidth]{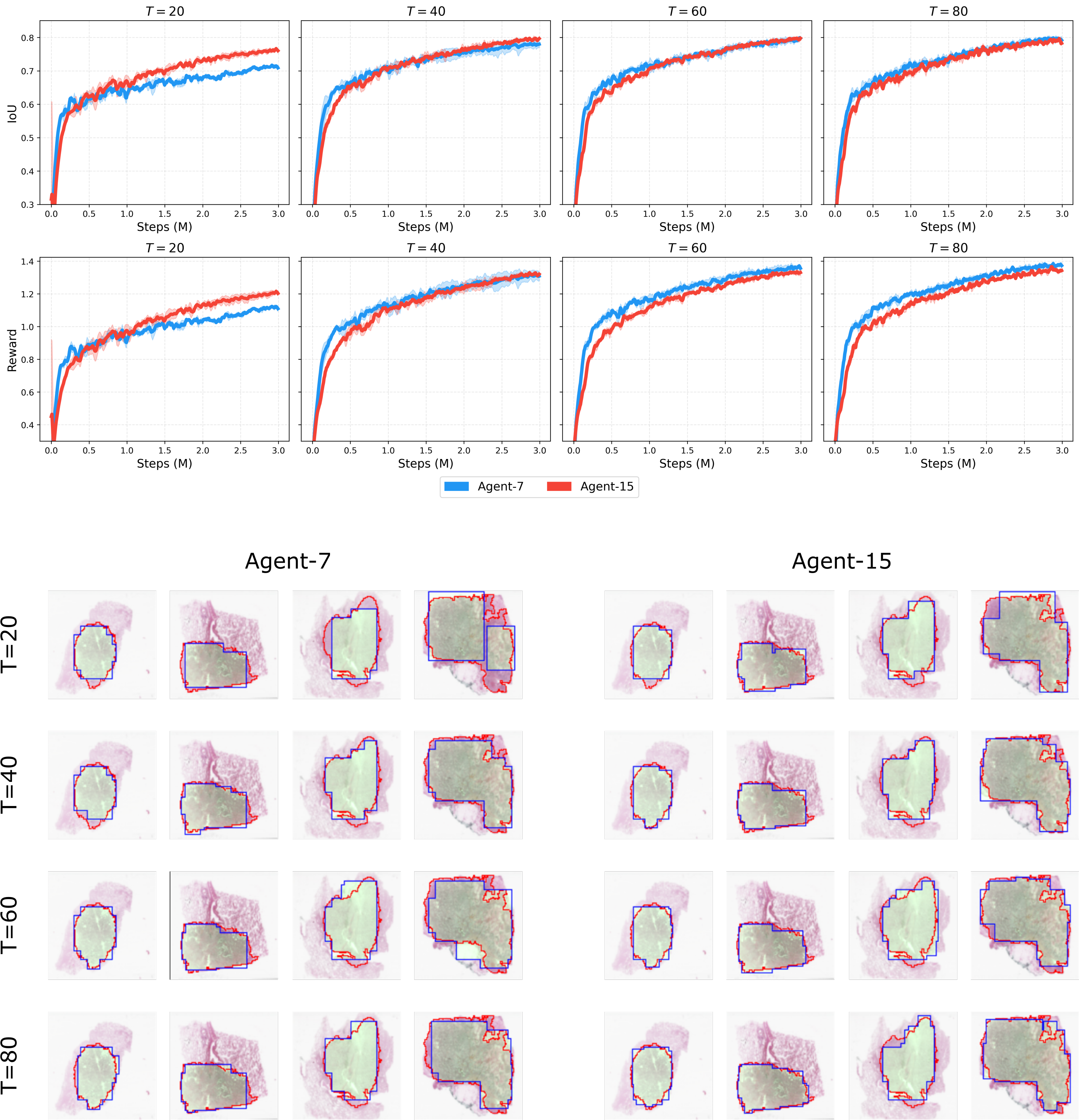}
\end{center}
   \caption{Effect of episode horizon $T$ and movement granularity on training dynamics and learned segmentations. Top: training IoU and reward curves for Agent-7 and Agent-15. Bottom: final segmentations masks obtained under different interaction budgets on training slides.}
\label{fig:Ts}
\end{figure}

Figure~\ref{fig:Ts} illustrates the effect of the episode horizon $T$ 
on both training dynamics and learned segmentation policies. Under 
constrained budgets ($T=20$), agents favour large coarse selections 
that rapidly increase tumour coverage while sacrificing boundary 
precision, producing visibly blockier masks. Increasing the step 
budget progressively shifts the learned policy towards finer boundary 
refinement, with smoother and more accurate final masks emerging at 
$T=60$ and $T=80$, these behaviours are consistent with the NSD improvement observed in Table~\ref{tab:T}.

The effect of movement granularity is most pronounced at low horizons. 
Under $T=20$, Agent-15 achieves substantially higher training IoU and 
reward than Agent-7, and qualitatively produces more refined 
segmentations. The additional movement options (e.g. m=1, m=0.5) enable more flexible spatial navigation within fewer steps, allowing a larger fraction of the episode budget to be spent on selection rather than movement. Agent-7, by contrast, compensates through larger coarse selections to maximise coverage under the same budget.

As $T$ increases, the performance gap between agents progressively 
decreases in both training curves and qualitative segmentation quality. 
Although Agent-15 provides finer navigation capability, its larger 
action space increases optimisation complexity, and finer displacements such as $m=0.1$ which could in principle enable more precise boundary delimitation but appears underutilised within the same training budget.

\label{appendix:steps}

\section{Hyperparameters and Implementation details}
\label{appendix:implementation}

Experiments use ImageNet-pretrained ResNet18 and RetCCL~\cite{retccl} pretrained ResNet50 together with dropout regularisation following~\cite{me}. Local and global image features are concatenated before being processed by independent actor and critic heads. For ResNet18, the concatenated representation is processed using two hidden layers of size 1024 in both the actor and critic networks. For ResNet50, a single hidden layer of size 4096 is used.

Training is performed using PPO with Adam optimisation and Generalized Advantage Estimation (GAE)~\cite{adam,GAE}. Linear scheduling is additionally applied to the learning rate, entropy coefficient, and PPO clipping parameter throughout training. Hyperparameter tuning in RL can significantly affect final performance. Table~\ref{tab:hyperparameters} summarises the main training and optimisation hyperparameters yielding the best results. Notably, a relatively high initial entropy coefficient was found to be important for encouraging sufficient early exploration in the WSI environment. For Agent-15-Res50 specifically, the training budget was increased from 6M to 8M steps as it requires more training.

Patch-based UNI baselines use frozen UNI feature representations extracted from image patches resized to $224\times224$. A lightweight two-layer MLP classifier is trained on top of the extracted features using cross-entropy loss and Adam optimisation with a learning rate of $10^{-5}$. Training for more than 6 epochs was found to overfit, and all classifiers were therefore trained for 4 epochs using balanced binary sampling and a 90/10 train--validation split. Final WSI segmentation masks are obtained through patch aggregation followed by connected-component filtering.

\begin{table*}[t]
\centering
\caption{Training and optimisation hyperparameters.}
\label{tab:hyperparameters}
\renewcommand{\arraystretch}{1.2}
\begin{tabular}{lcc}
\hline
Hyperparameter      & extended-cohort & original-cohort \\
\hline
Episode horizon ($T$)        & 80 & 80   \\
Observation size ($p$)       & 112& 112 \\
Parallel environments        & 16 & 16   \\
Rollout steps per environment   & 1000 & 500  \\
Discount factor $\gamma$     & 0.99 & 0.99 \\
GAE parameter $\lambda$      & 0.95 & 0.95 \\
Value loss coefficient $c_1$    & 1.0 & 1.0    \\
Entropy coefficient $c_2$       & $4\times10^{-2}$ & $4\times10^{-2}$  \\
Batch size                   & 256  & 128   \\
Learning rate                & $4\times10^{-5}$  & $2\times10^{-5}$ \\
clip epsilon $\epsilon$                & 0.2  &  0.2 \\
Weight decay                 & $1\times10^{-4}$  & $1\times10^{-4}$ \\
Gradient clipping            & 1.0   & 1.0   \\
Dropout rate                 & 0.10  & 0.10 \\
Advantage normalization      & Yes   & Yes \\
Input normalization (ImageNet mean/std) & No  & No  \\
Reward normalization                     & No   & No   \\
PPO epochs                               & 4 & 4 \\
Total training steps                     & 6M & 3M \\
MDR ($\alpha_1,\alpha_2$)   & (2,2) & (2,2) \\
\hline
\end{tabular}
\end{table*}

\section{Inference time analysis details}
\label{appendix:time}
To ensure a fair comparison, the inference time analysis for all models was conducted using the same hardware configuration. Specifically, each experiment was executed on a single NVIDIA Tesla V100 GPU (32 GB VRAM), 6 CPU cores of an Intel Xeon Gold 6126 processor, and 384 GB of system memory. 

The best-case and worst-case scenarios correspond to WSIs requiring the shortest and longest processing times, respectively. For patch-based methods, processing time is primarily influenced by slide dimensions, as larger slides generate a greater number of patches that must be analyzed. For agent-based methods, processing time is additionally affected by the slide size through its impact on the image pyramid structure, including the pyramid depth and the availability of image representations at different downsampling levels. Consequently, larger WSIs generally require more computational effort and longer processing times.

\end{document}